\documentclass[conference]{IEEEtran}
\usepackage{cite}

\ifCLASSINFOpdf
   \usepackage[pdftex]{graphicx}
\else
   \usepackage[dvips]{graphicx}
\fi
\usepackage{amsmath}
\usepackage{algorithmic}
\usepackage{algorithm}

\usepackage{array}

\ifCLASSOPTIONcompsoc
  \usepackage[caption=false,font=normalsize,labelfont=sf,textfont=sf]{subfig}
\else
  \usepackage[caption=false,font=footnotesize]{subfig}
\fi
\usepackage{url}

\mathchardef\mhyphen="2D 
\usepackage{color,soul}
\usepackage{dsfont}
\usepackage[utf8]{inputenc}
\usepackage{float}
\usepackage{booktabs}
\DeclareMathOperator{\logit}{logit}
\DeclareMathOperator{\xentropy}{H}
\DeclareMathOperator{\clamp}{clamp}

\begin{document}
%

\title{Exploring the Space of Adversarial Images}


\author{
	Pedro Tabacof and Eduardo Valle \\
	RECOD Lab. --- DCA / School of Electrical and Computer Engineering (FEEC) \\
	University of Campinas (Unicamp) \\
	Campinas, SP, Brazil \\
	\{tabacof, dovalle\}@dca.fee.unicamp.br 
	}


\IEEEoverridecommandlockouts
\IEEEpubid{\makebox[\columnwidth]{
		\copyright 2016 IEEE -- Manuscript accepted at IJCNN 2016\hfill} \hspace{\columnsep}\makebox[\columnwidth]{ }}

\maketitle

\begin{abstract}
Adversarial examples have raised questions regarding the robustness and security of deep neural networks. In this work we formalize the problem of adversarial images given a pretrained classifier, showing that even in the linear case the resulting optimization problem is nonconvex. We generate adversarial images using shallow and deep classifiers on the MNIST and ImageNet datasets. We probe the pixel space of adversarial images using noise of varying intensity and distribution. We bring novel visualizations that showcase the phenomenon and its high variability. We show that adversarial images appear in large regions in the pixel space, but that, for the same task, a shallow classifier seems more robust to adversarial images than a deep convolutional network.
\end{abstract}


%
\IEEEpeerreviewmaketitle

\section{Introduction}

Small but purposeful pixel distortions can easily fool the best deep convolutional networks for image classification \cite{szegedy2013intriguing, nguyen2014deep}. The small distortions are hardly visible by humans, but still can mislead most neural networks. Adversarial images can even fool the internal representation of images by neural networks \cite{sabour2015adversarial}. That problem has divided the Machine Learning community, with some hailing it as a ``deep flaw'' of deep neural networks \cite{deepflaw}; and others promoting a more cautious interpretation, and showing, for example, that most classifiers are susceptible to adversarial examples \cite{goodfellow2014explaining, fawzi2015analysis}.  

The distortions were originally obtained via an optimization procedure, \cite{szegedy2013intriguing}, but it was subsequently shown that adversarial images could be generated using a simple gradient calculation \cite{goodfellow2014explaining} or using evolutionary algorithms \cite{nguyen2014deep}. 

Despite the controversy, adversarial images surely suggest a lack of robustness, since they are (for humans) essentially equal to correctly classified images. Immunizing a network against those perturbations increases its ability to generalize, a form of regularization \cite{goodfellow2014explaining} whose statistical nature deserves further investigation. Even the traditional backpropagation training procedure can be improved with adversarial gradient \cite{nokland2015improving}. Recent work shows that it is possible for the training procedure to make classifiers robust to adversarial examples by using a strong adversary \cite{huang2015learning} or defensive distillation \cite{papernot2015distillation}. 

\begin{figure}[h!]
	\centering
	\includegraphics[width=.99\linewidth]{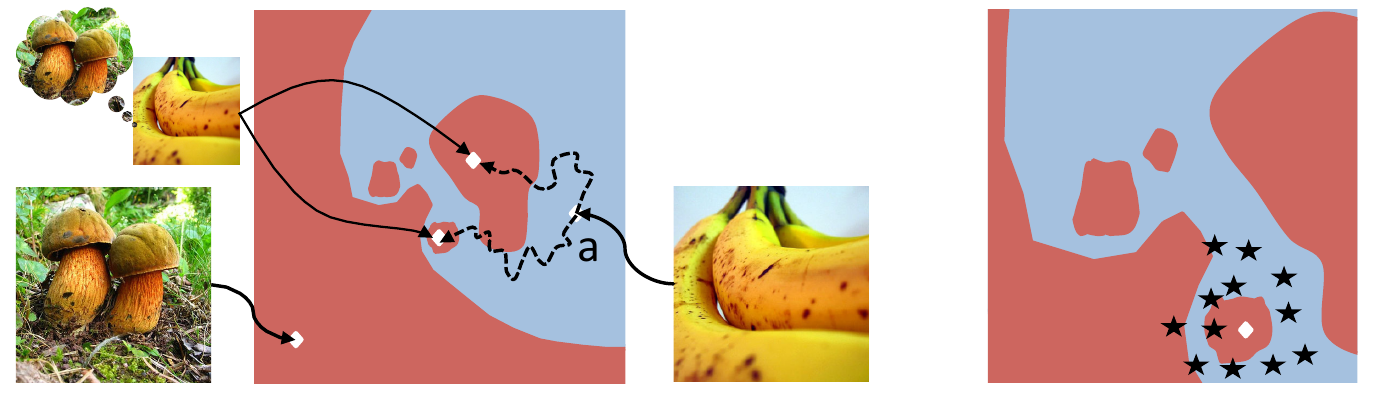}
	\caption{Fixed-sized images occupy a high-dimensional space spanned by their pixels (one pixel = one dimension), here depicted as a 2D colormap. \textbf{Left:} classifiers associate points of the input pixel space to output class labels, here `banana' (blue) and `mushroom' (red). From a correctly classified original image (a), an optimization procedure (dashed arrows) can find adversarial examples that are, for humans, essentially equal to the original, but that will fool the classifier. \textbf{Right:} we probe the pixel space by taking a departing image (white diamond), adding random noise to it (black stars), and asking the classifier for the label. In compact, stable regions, the classifier will be consistent (even if wrong). In isolated, unstable regions, as depicted, the classifier will be erratic.}
	\label{fig:graphical-abs}
\end{figure}

In this paper, we extend previous works on adversarial images for deep neural networks \cite{szegedy2013intriguing}, by exploring the pixel space of such images using random perturbations. That framework (Figure~\ref{fig:graphical-abs}) allows us to ask interesting questions about adversarial images. Initial skepticism about the relevance of adversarial images suggested they existed as isolated points in the pixel space, reachable only by a guided procedure with complete access to the model. More recent works \cite{ goodfellow2014explaining, gu2014towards} claim that they inhabit large and contiguous regions in the space. The correct answer has practical implications: if adversarial images are isolated or inhabit very thin pockets, they deserve much less worry than if they form large, compact regions. In this work we intend to shed light to the issue with an in-depth analysis of adversarial image space.

\section{Creating adversarial images}
Assume we have a pre-trained classifier $p=f(X)$ that, for each input $X\in\mathcal{I}$, corresponding to the pixels of a fixed-sized image, outputs a vector of probabilities $p=[p_1 \cdots p_i \cdots p_n]$ of the image belonging to the class label $i$. 
We can assign $h$ to the label corresponding to the highest probability $p_h$. Assume further that $\mathcal{I}=[L-U]$, for grayscale images, or $\mathcal{I}=[L-U]^3$ for RGB images, where $L$ and $U$ are the lower and upper limits of the pixel scale. 

Assume that $c$ is the correct label and that we start with $h=c$, otherwise there is no point in fooling the classifier. We want to add the smallest distortion $D$ to $X$, such that the highest probability will no longer be assigned to $h$. The distortions must keep the input inside its space, i.e., we must ensure that $X+D\in\mathcal{I}$. In other words, the input is box-constrained. Thus, we have the following optimization:

\begin{equation}
\begin{aligned}
& \underset{D}{\text{minimize}}
& & \| D \| \\
& \text{subject to}
& & L \leq X + D \leq U \\
& & & p = f(X+D) \\
& & & \max(p_1 - p_c, ..., p_n - p_c) > 0
\end{aligned}
\label{eq:orig}
\end{equation}

That formulation is more general than the one presented by \cite{szegedy2013intriguing}, for it ignores non-essential details, such as the choice of the adversarial label. It also showcases the non-convexity: since $\max(x) < 0$ is convex, the inequality is clearly concave \cite{boyd2004convex}, making the problem non-trivial even if the model $p=f(X)$ were linear in $X$.  
Deep networks, of course, exacerbate the non-convexity due to their highly non-linear model. For example, a simple multi-layer perceptron for binary classification could have $f(X)=\logit^{-1}(W_2 \cdot \tanh(W_1 \cdot X + b_1) + b_2)$, which is neither convex nor concave due to the hyperbolic tangent.

\subsection{Procedure}
\label{sec:procedure}

Training a classifier usually means minimizing the classification error by changing the model weights. To generate adversarial images, however, we hold the weights fixed, and find the minimal distortion that still fools the network. 

We can simplify the optimization problem of eq.~\ref{eq:orig} by exchanging the $\max$ inequality for a term in the loss function that measures how adversarial the probability output is:

\begin{equation}
\begin{aligned}
& \underset{D}{\text{minimize}}
& & \| D \| + C\cdot\xentropy(p, p^A)\\
& \text{subject to}
& & L \leq X + D \leq U \\
& & & p = f(X+D) \\
\end{aligned}
\label{eq:min}
\end{equation}

where we introduce the adversarial probability target $p^A=[\mathds{1}_{i=a}]$, which assigns zero probability to all but a chosen adversarial label $a$. That formulation is essentially the same of \cite{szegedy2013intriguing}, picking an explicit (but arbitrary) adversary label. We stipulate the loss function: the cross-entropy ($\xentropy$) between the probability assignments; while \cite{szegedy2013intriguing} keep that choice open.

The constant $C$ balances the importance of the two objectives. The lower the constant, the more we will  minimize the distortion norm. Values too low, however, may turn the optimization unfeasible. We want the lowest, but still feasible, value for $C$.

We can solve the new formulation with any local search compatible with box-constraints. Since the optimization variables are the pixel distortions, the problem size is exactly the size of the network input, which in our case varies from $28\times28=784$ for MNIST \cite{lecun1998mnist} to $221\times221\times3=146\,523$ for OverFeat \cite{sermanet2013overfeat}.
In contrast to current neural network training, that reaches hundreds of millions of weights, those sizes are small enough to allow second-order procedures, which converge faster and with better guarantees \cite{nocedal2006numerical}. We chose L-BFGS-B, a box-constrained version of the popular L-BFGS second-order optimizer \cite{zhu1997algorithm}. We set the number of corrections in the limited-memory matrix to 15, and the maximum number of iterations to 150. We used Torch7 to model the networks and extract their gradient with respect to the inputs
\cite{collobert2011torch7}. Finally, we implemented a bisection search to determine the optimal value for $C$ \cite{kaw2009numerical}. The algorithm is explained in detail in the next section.

\subsection{Algorithm}

Algorithm~\ref{alg} implements the optimization procedure used to find the adversarial images. The algorithm is essentially a bisection search for the constant $C$, where in each step we solve a problem equivalent to Eq.~\ref{eq:min}. Bisection requires initial lower and upper bounds for $C$, such that the upper bound succeeds in finding an adversarial image, and the lower bound fails. It will then search the transition point from failure to success (the ``zero'' in a root-finding sense): that will be the best $C$. We can use $C=0$ as lower bound, as it always leads to failure (the distortion will go to zero). To find an upper bound leading to success, we start from a very low value, and exponentially increase it until we succeed. During the search for the optimal $C$ we use warm-starting in L-BFGS-B to speed up convergence: the previous optimal value found for $D$ is used as initial value for the next attempt.

To achieve the general formalism of eq.~\ref{eq:orig} we would have to find the adversarial label leading to minimal distortion. However, in datasets like ImageNet \cite{deng2009imagenet}, with hundreds of classes, this search would be too costly. Instead, in our experiments, we opt to consider the adversarial label as one of the sources of random variability. As we will show, this does not upset the analyses.

The source code for adversarial image generation and pixel space analysis can be found in \url{https://github.com/tabacof/adversarial}.

\newcommand\lbfgsb{\mathop{L\mhyphen BFGS\mhyphen B}}
\begin{algorithm}
	\caption{Adversarial image generation algorithm}
	\begin{algorithmic}[1]
	\label{alg}
	\REQUIRE A small positive value $\epsilon$ 
	\ENSURE $\lbfgsb(X, p^A, C)$ solves optimization \ref{eq:min}
	\STATE \COMMENT{Finding initial $C$}
	\STATE $C \gets \epsilon$ 
	\REPEAT
		\STATE $C \gets 2\times C$
		\STATE $D, p \gets \lbfgsb(X, p^A, C)$
	\UNTIL {$\max(p_i)$ in $p$ is $p_a$}
	\STATE \COMMENT {Bisection search}
	\STATE $C_{low} \gets 0$, $C_{high} \gets C$ 
	\REPEAT 
		\STATE $ C_{half} \gets (C_{high} +C_{low})/2 $
		\STATE $D', p \gets \lbfgsb(X, p^A, C_{half})$ 
		\IF {$\max(p_i)$ in $p$ is $p_a$} 
			\STATE $D \gets D'$
			\STATE $C_{high} \gets C_{half}$
		\ELSE 
			\STATE $C_{low} \gets C_{half}$
		\ENDIF
	\UNTIL {$(C_{high}-C_{low}) < \epsilon$} 
	\RETURN $D$ \
	\end{algorithmic}
\end{algorithm}
	
\section{Adversarial space exploration}

\begin{figure*}[!t]
	\centering
	\subfloat[MNIST with logistic regression. The correct labels are self-evident.] {
		\includegraphics[width=6in]{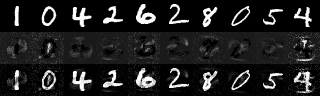}
		\label{fig:ex_mnist} }
	\hfil
	\subfloat[MNIST with convolutional network. The correct labels are self-evident.] {
		\includegraphics[width=6in]{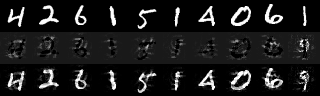}
		\label{fig:ex_imagenet_mnist_conv} }
	\hfil
	\subfloat[OverFeat on ImageNet. From left to right, correct labels: `Abaya', `Ambulance', `Banana', `Kit Fox', `Volcano'. Adversarial labels for all: `Bolete' (a type of mushroom).] {
		\includegraphics[width=6in]{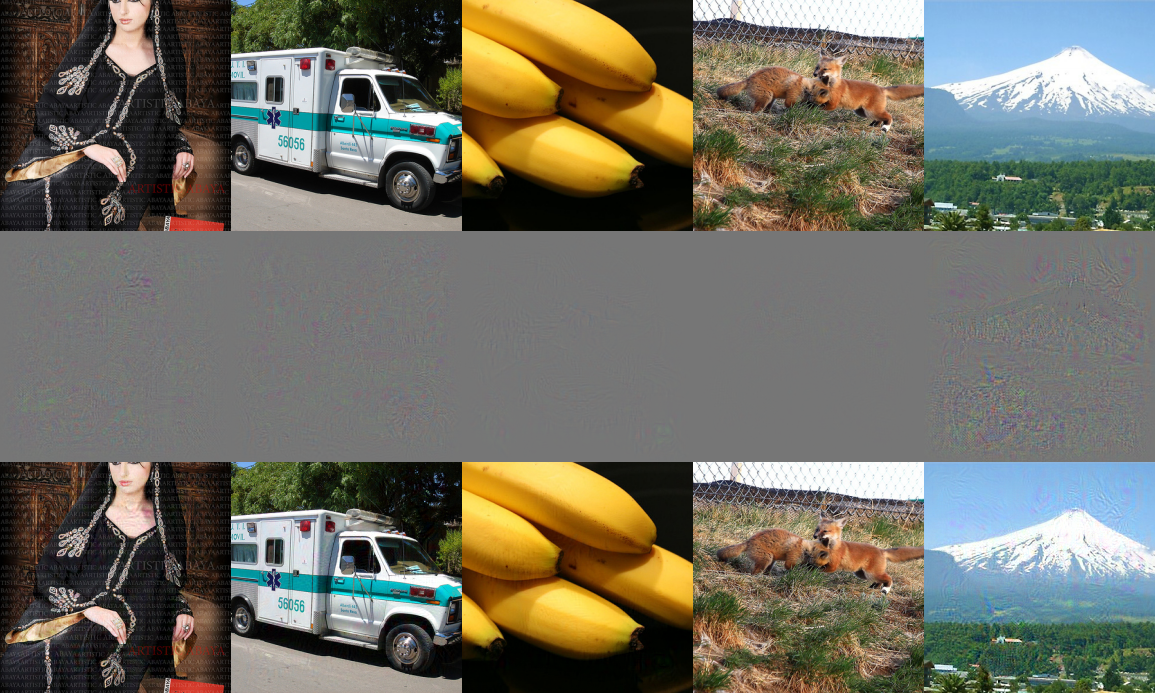}
		\label{fig:ex_imagenet} }
	
	\caption{Adversarial examples for each network. For all experiments: original images on the top row, adversarial images on the bottom row, distortions (difference between original and adversarial images) on the middle row.}
	\label{fig:ex_imagenet_mnist}
\end{figure*}

In this section we explore the vector space spanned by the pixels of the images to investigate the ``geometry'' of adversarial images: are they isolated, or do they exist in dense, compact regions? Most researchers currently believe that images of a certain appearance (and even meaning) are contained into relatively low-dimensional manifolds inside the whole space \cite{bengio2009learning}. However, those manifolds are exceedingly convoluted, discouraging direct geometric approaches to investigate the pixel space.

Thus, our approach is indirect, probing the space around the images with small random perturbations. In regions where the manifold is nice --- round, compact, occupying most of the space --- the classifier will be consistent (even if wrong). In the regions where the manifold is problematic --- sparse, discontinuous, occupying small fluctuating subspaces --- the classifier will be erratic.

\subsection{Datasets and models}

To allow comparison with the results of \cite{szegedy2013intriguing}, we employ the MNIST handwritten digits database (10 classes, 60k training and 10k testing images), and the 2012 ImageNet Large Visual Recognition Challenge Dataset (1000 classes, 1.2M+ training and 150k testing images).

For MNIST, \cite{szegedy2013intriguing} tested convolutional networks and autoencoders. We employ both convolutional networks and a logistic linear classifier. While logistic classifiers have limited accuracy ($\sim$7.5\% error), their training procedure is convex \cite{boyd2004convex}. They also allowed us to complement the original results \cite{szegedy2013intriguing} by investigating adversarial images in a shallow classifier.

The convolutional network we employed for MNIST/ConvNet consisted of two convolutional layers with 64 and 128 5$\times$5 filters, two 2$\times$2 max-pooling layers after the convolutional layers, one fully-connected layer with 256 units, and a softmax layer as output. We used ReLU for nonlinearity, and 0.5 dropout before the two last layers. The network was trained with SGD and momentum. Without data augmentation, this model achieves 0.8\% error on the test set.

For ImageNet, we used the pre-trained OverFeat network \cite{sermanet2013overfeat}, which achieved 4th place at the ImageNet competition in 2013, with 14.2\% top-5 error in the classification task, and won the localization competition the same year. \cite{szegedy2013intriguing} employed AlexNet \cite{krizhevsky2012imagenet}, which achieved 1st place at the ImageNet competition in 2012, with 15.3\% top-5 error.  

On each dataset, we preprocess the inputs by standardizing each pixel with the global mean and standard deviation of all pixels in the training set images.



Figure~\ref{fig:ex_imagenet_mnist} illustrates all three cases. Original and adversarial images are virtually indistinguishable. The pixel differences (middle row) do not show any obvious form --- although a faint ``erasing-and-rewriting'' effect can be observed for MNIST.  
Figures \ref{fig:ex_mnist} and \ref{fig:ex_imagenet_mnist_conv} also suggest that the MNIST classifiers are more robust to adversarial images, since the distortions are larger and more visible. 
We will see, throughout the paper, that classifiers for MNIST and for ImageNet have important differences in how they react to adversarial images.

\subsection{Methods}
\label{sec:methods}

Each case (MNIST/Logistic, MNIST/ConvNet, ImageNet/OverFeat) was investigated independently, by applying the optimization procedure explained in Section~\ref{sec:procedure}. For ImageNet we sampled 5 classes (Abaya, Ambulance, Banana, Kit Fox, and Volcano), 5 correctly classified examples from each class, and sampled 5 adversarial labels (Schooner, Bolete, Hook, Lemur, Safe), totaling 125 adversarial images. For MNIST, we just sampled 125 correctly classified examples from the 10K examples in the test set, and sampled an adversarial label (from 9 possibilities) for each one. All random sampling was made with uniform probability. To sample only correctly classified examples, we rejected the  misclassified ones until we accumulated the needed amount. We call, in the following sections, those correctly classified images \emph{originals}, since the adversarial images are created from them.

The probing procedure consisted in picking an image pair (an adversarial image and its original), adding varying levels of noise to their pixels, resubmitting both to the classifier, and observing if the newly assigned labels corresponded to the original class, to the adversarial class, or to some other class. 

We measured the \emph{levels of noise} ($\lambda$) relative to the difference between each image pair. We initially tested a Gaussian i.i.d. model for the noise. For each image $X = \{x_i\}$, our procedure creates an image $X' = \{ \clamp(x_i + \epsilon) \}$ where $\epsilon \sim \mathcal{N}(\mu, \lambda\sigma^2)$, and $\mu$ and $\sigma^2$ are the sample mean and variance of the distortion pixels. In the experiments we ranged $\lambda$ from $2^{-5}$ to $2^5$. To keep the pixel values of $X'$ within the original range $[L-U]$ we employ $\clamp(x)=\min(\max(x,L),U)$. In practice, we observed that clamping has little effect on the noise statistics.

An i.i.d. Gaussian model discards two important attributes of the distortions: spatial correlations, and  higher-order momenta. We wanted to evaluate the relative importance of those, and thus performed an extra round of experiments that, while still discarding all spatial correlations by keeping the noise i.i.d., adds higher momenta information by modeling non-parametrically the distribution of distortion pixels. Indeed, a study of those higher momenta (Table~\ref{table:momenta}) suggests that the adversarial distortions has a much heavier tail than the Gaussians, and we wanted to investigate how that affects the probing. The procedure is exactly the same as before, but with $\epsilon \sim \mathcal{M}$, where $\mathcal{M}$ is an empirical distribution induced by a non-parametric observation of the distortion pixels. In those experiments we cannot control the level: the variance of the noise is essentially the same as the variance of the distortion pixels.

Our main metric is the fraction of images (in \%) that keep or switch labels when noise is added to a departing image, which we use as a measure of the stability of the classifier at the departing image in the pixel space. The fraction is computed over a sample of 100 probes, each probe being a repetition of the experiment with all factors held fixed but the sampling of the random noise.

\begin{table}[!t]
	\renewcommand{\arraystretch}{1.3}
	\centering
	\caption{Descriptive statistics of the adversarial distortions for the two datasets averaged over the 125 adversarial examples. Pixels values in $[0-255]$. Logistic and ConvNet refer to MNIST dataset, OverFeat refers to ImageNet dataset.
	}
	\begin{tabular}{*5c}
		\toprule
		& Mean  & S.D. & Skewness & Ex. Kurtosis \\
		\hline \\
		Logistic    & $30.7 \pm 4.3$ & $18.3 \pm 11.3$ & $0.1 \pm 1.0$ & $7.8\pm 3.2$ \\
		ConvNet     & $27.5 \pm 2.1$ & $23.0 \pm 9.4$ & $-0.5 \pm 1.6$ & $17.6\pm 7.3$ \\
		OverFeat & $118.4 \pm 0.1$ & $ 1.9 \pm 2.0$ & $0.0 \pm 0.1$ & $6.5\pm 4.1$ \\
		\bottomrule
	\end{tabular}
	\label{table:momenta}
\end{table}

\subsection{Results}

\begin{figure*}[t!]
	\centering
	\subfloat[] {
		\includegraphics[width=3in]{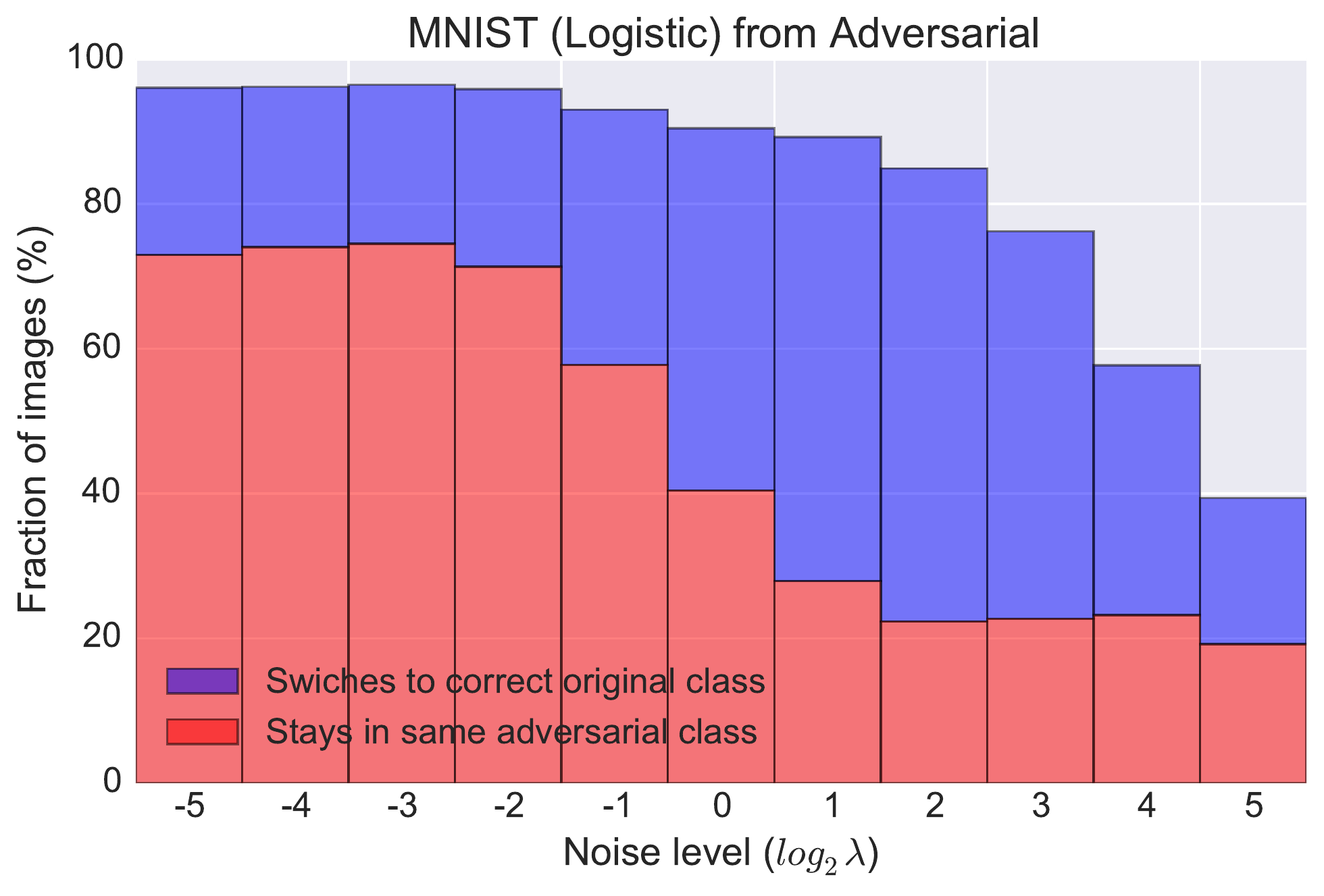}
		\label{fig:mnist_from_adv} }
	\subfloat[] {
		\includegraphics[width=3in]{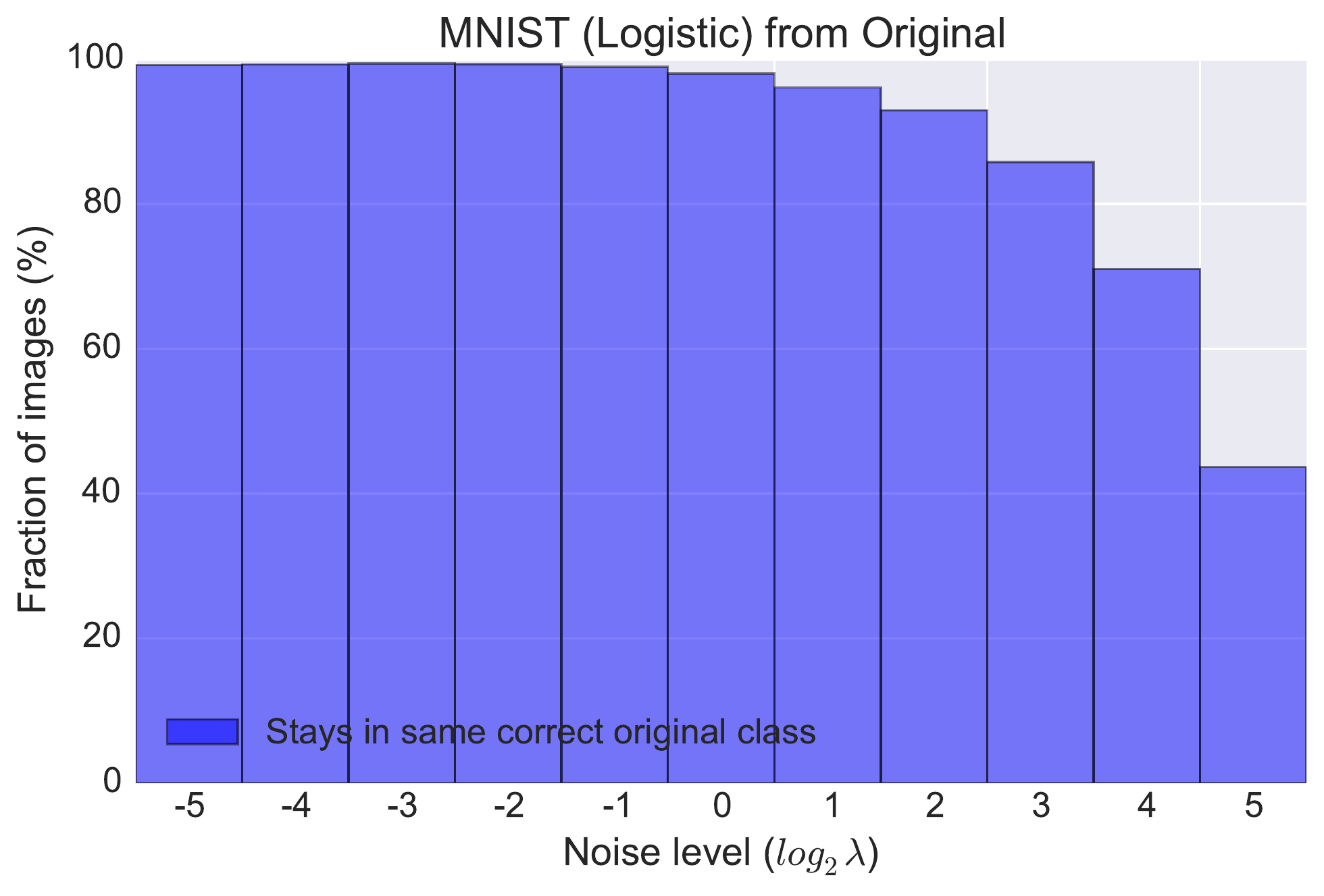}
		\label{fig:mnist_from_orig} }
	\hfil
	\subfloat[] {
		\includegraphics[width=3in]{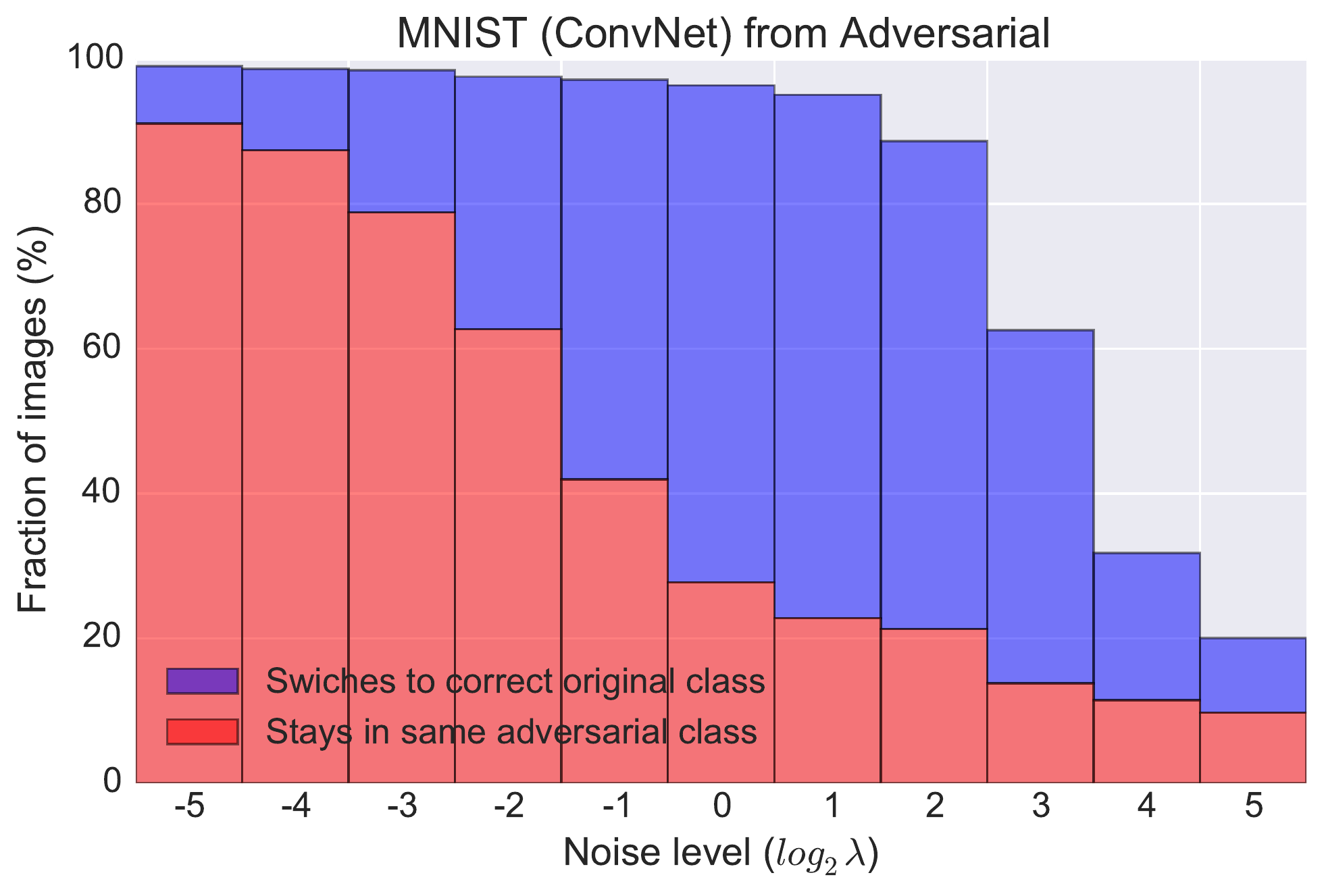}
		\label{fig:mnist_conv_from_adv} }
	\subfloat[]{
		\includegraphics[width=3in]{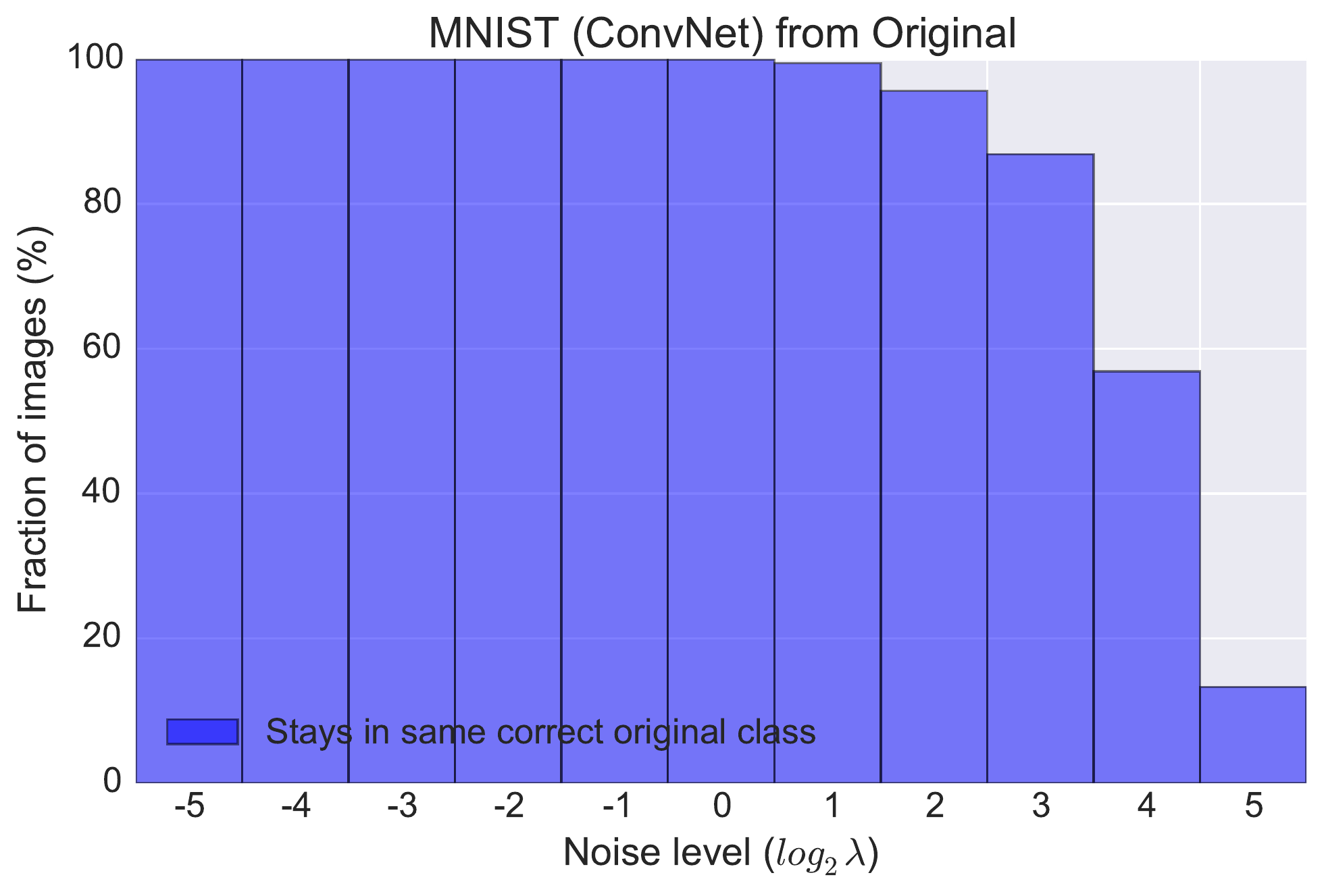}
		\label{fig:mnist_conv_from_orig} }
	\hfil
	\subfloat[]{
		\includegraphics[width=3in]{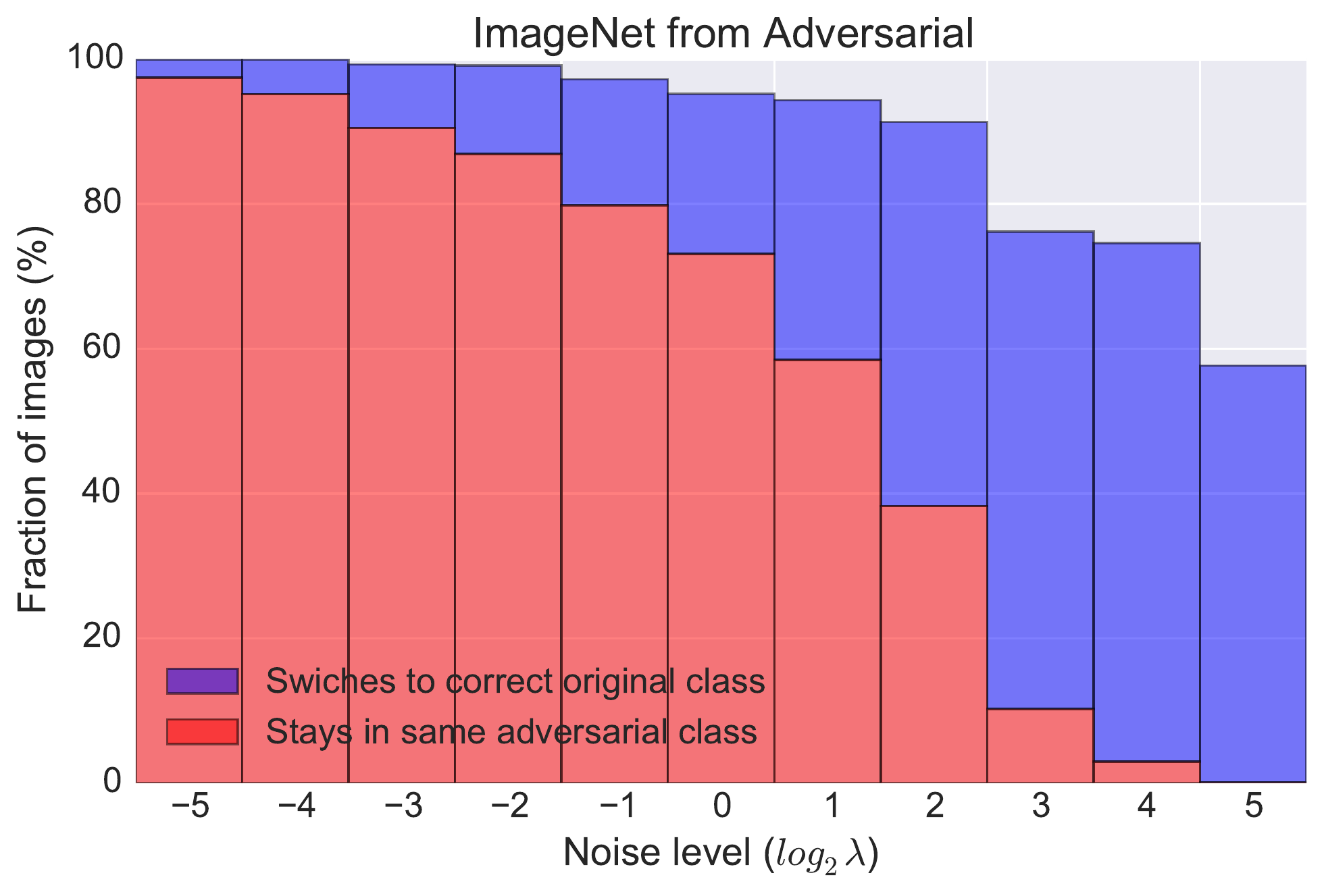}
		\label{fig:from_adv} }
	\subfloat[]{
		\includegraphics[width=3in]{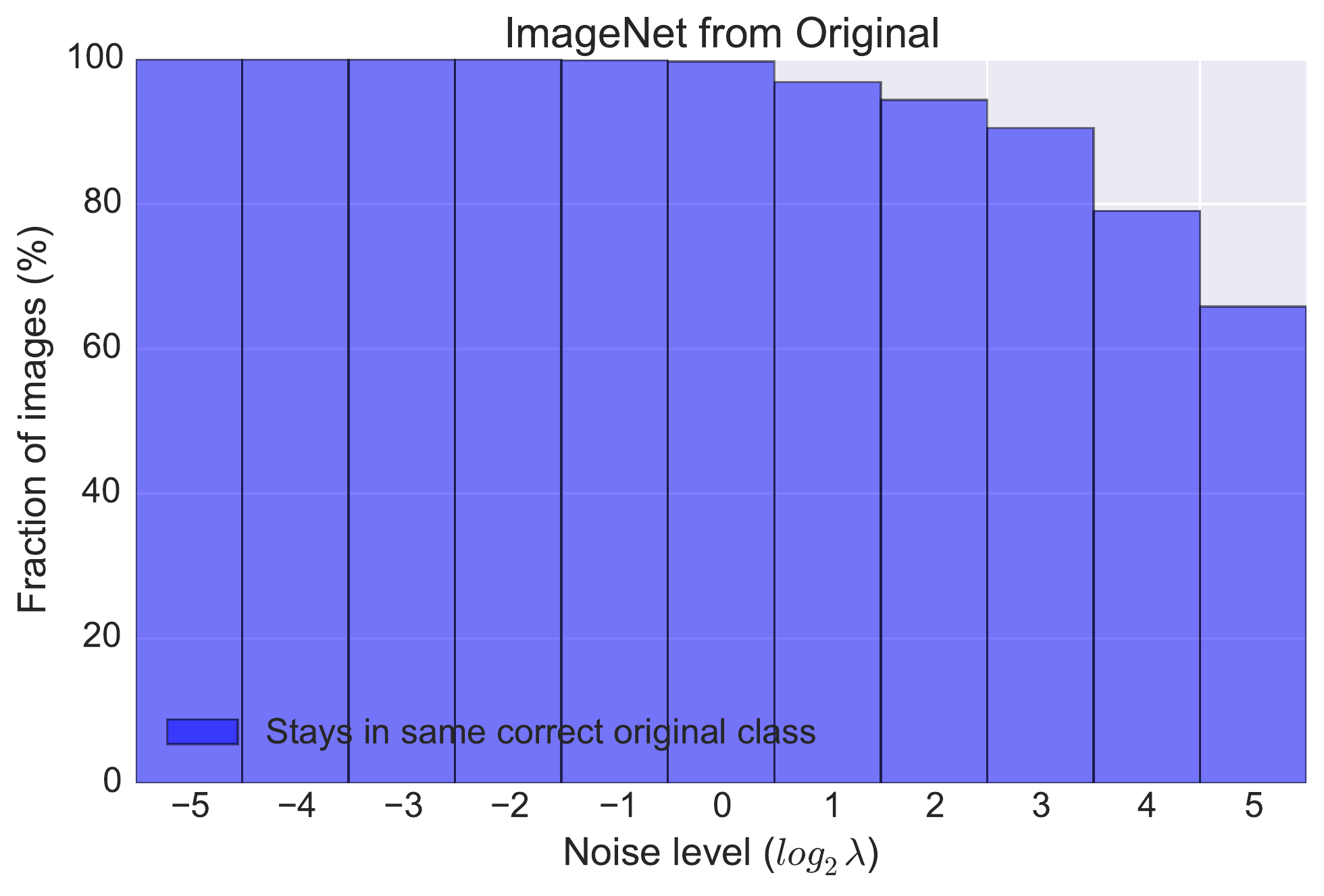}
		\label{fig:from_orig} }
	
	\caption{Adding Gaussian noise to the images. We perform the probing procedure explained in Section~\ref{sec:methods} to measure the stability of the classifier boundaries at different points of the pixel space. To escape the adversarial pockets completely we have to add a noise considerably stronger than the original distortion used to reach them in the first place: adversarial regions are not isolated. That is especially true for ImageNet/OverFeat. Still, the region around the correctly classified original image is much more stable. This graph is heavily averaged: each stacked column along the horizontal axis summarizes 125 experiments $\times$ 100 random probes.}
	\label{fig:all_averages}
\end{figure*}

Figure \ref{fig:all_averages} shows that adversarial images do not appear isolated. On the contrary, to completely escape the adversarial pocket we need to add a noise with much higher variance --- notice that the horizontal axis is logarithmic --- than the distortion used to reach the adversarial image in the first place. 


In both networks, the original images display a remarkable robustness against Gaussian noise (Figures~\ref{fig:mnist_from_orig} and~\ref{fig:from_orig}), confirming that robustness to random noise does not imply robustness to adversarial examples \cite{fawzi2015analysis}. That shows that while the adversarial pockets are not exactly isolated, neither are they as well-behaved as the zones that contain the correctly classified samples.





\begin{figure*}[!t]
	\centering
	\subfloat [] {
		\includegraphics[width=3in]{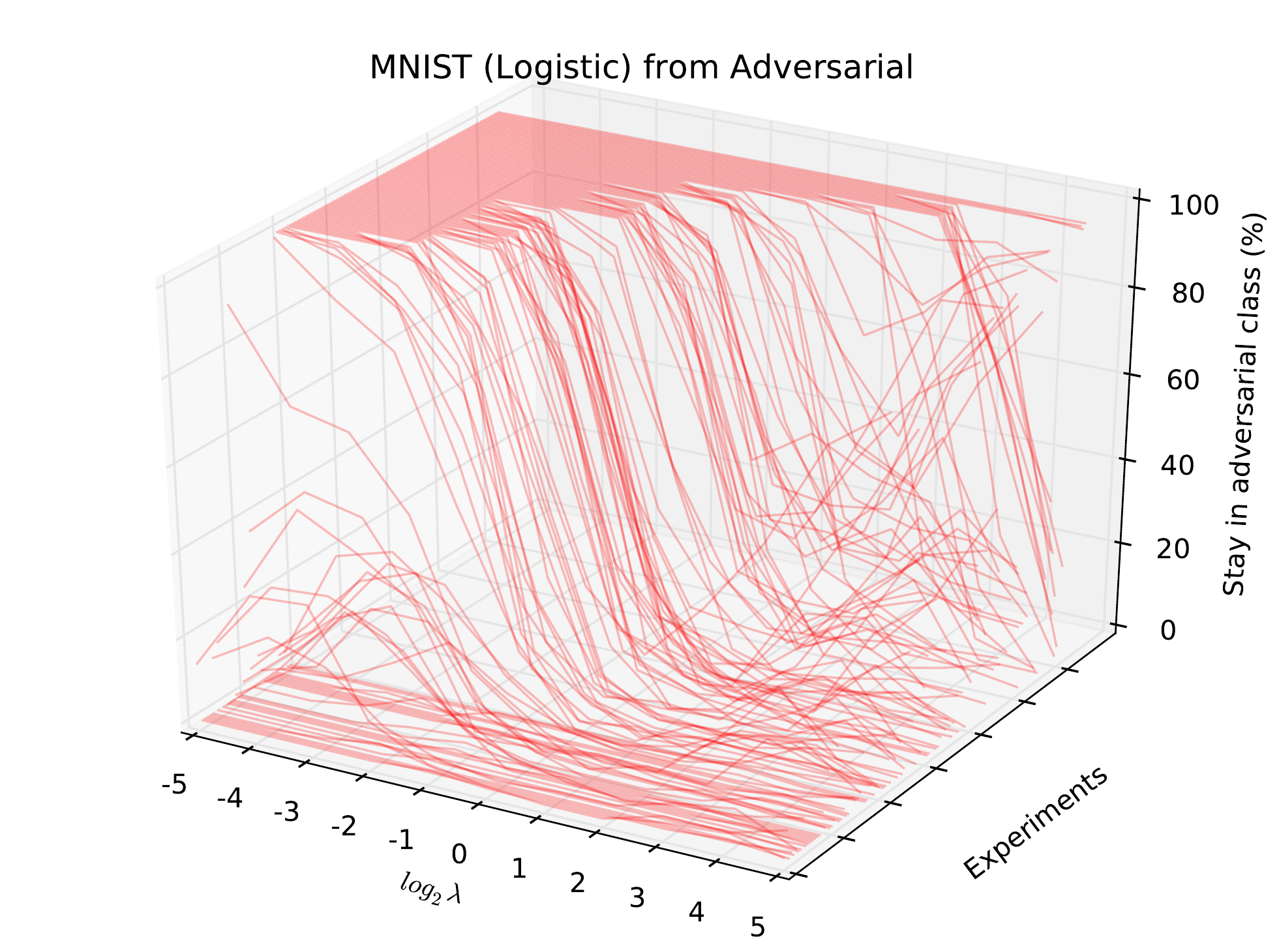}
		\label{fig:mnist_3d_adv_from_adv} }
	\subfloat [] {
		\includegraphics[width=3in]{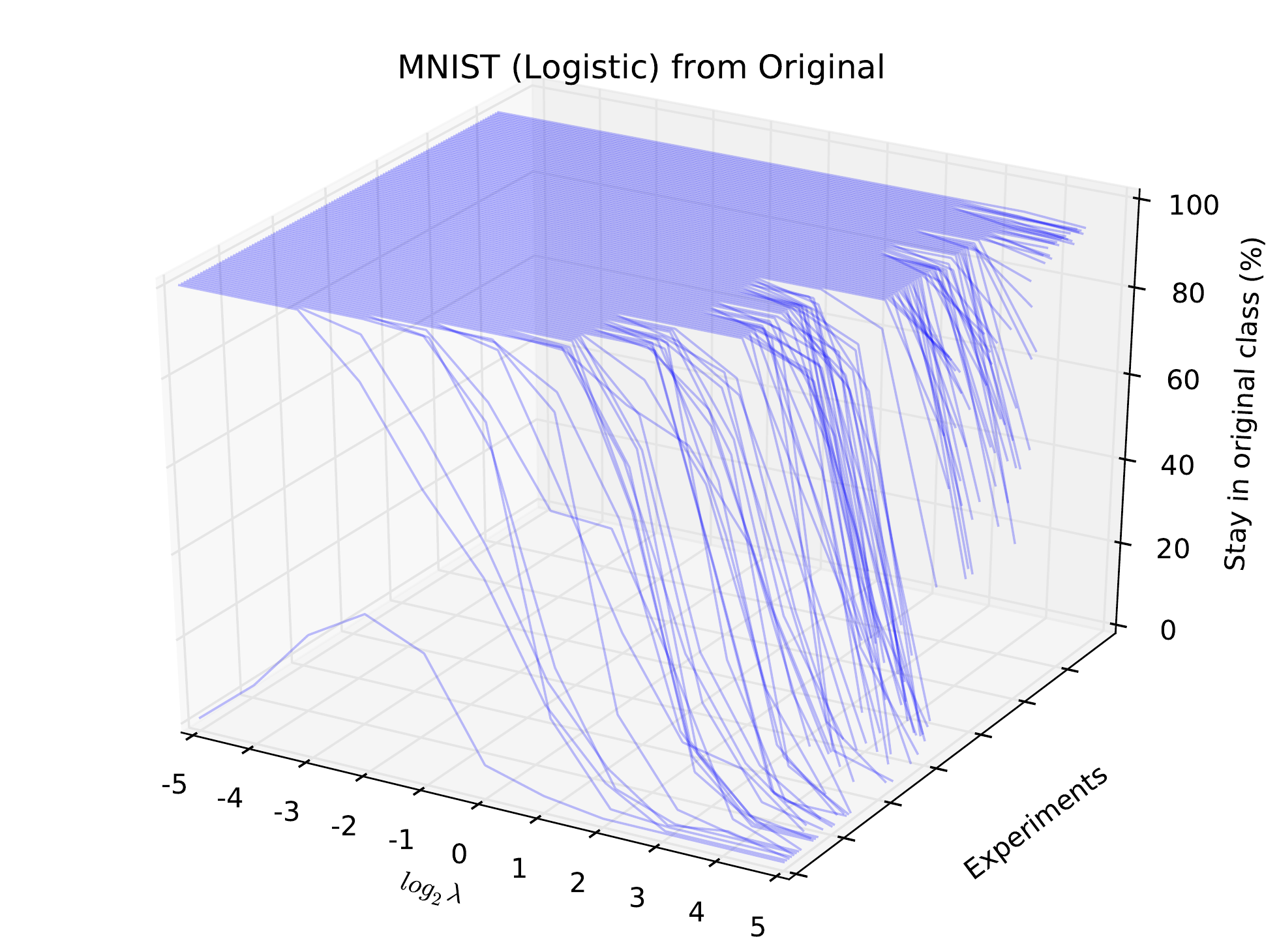}
		\label{fig:mnist_3d_orig_from_orig} }
	\hfil
	\subfloat [] {
		\includegraphics[width=3in]{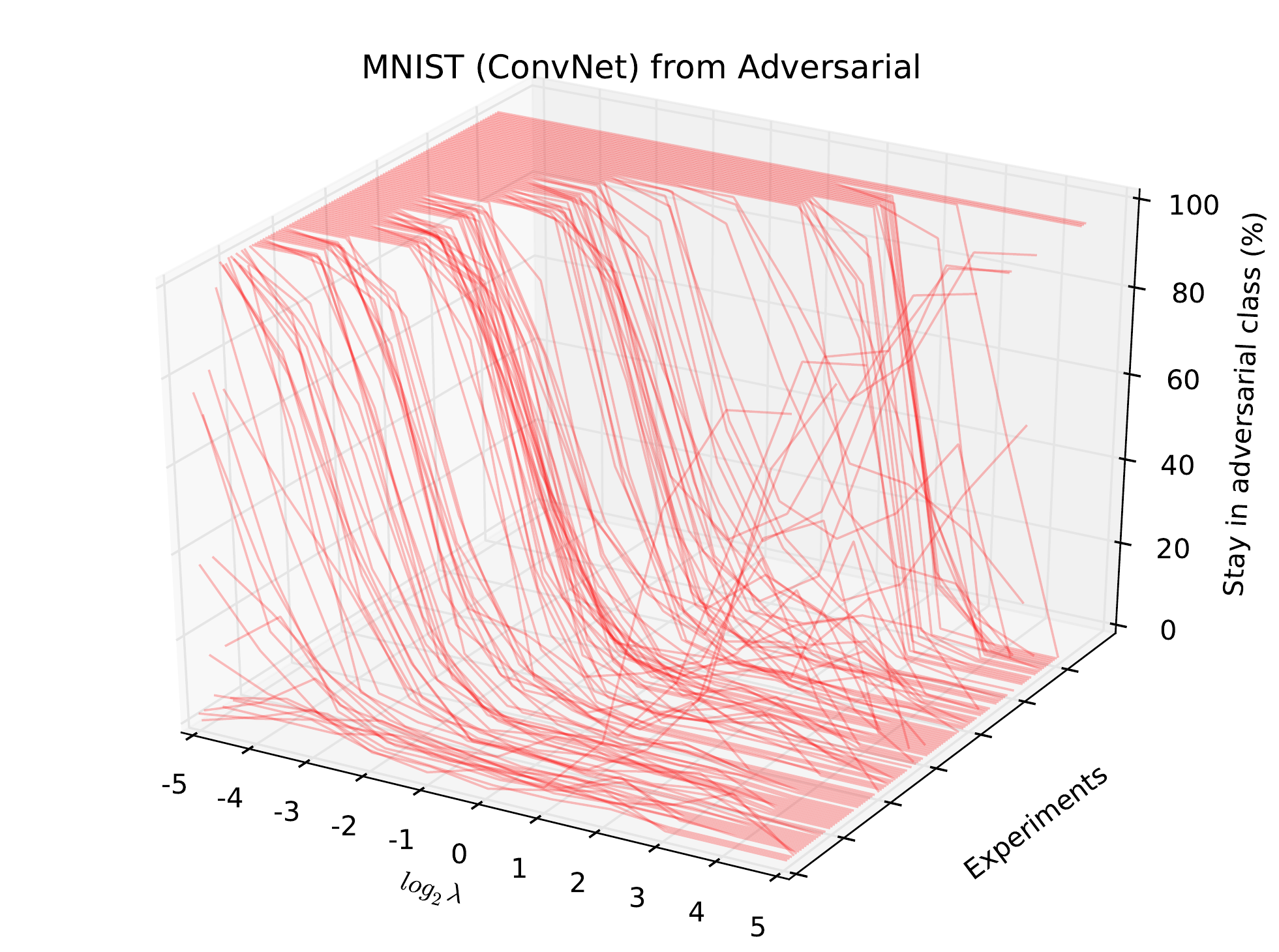}
		\label{fig:mnist_conv_3d_adv_from_adv} }
	\subfloat [] {
		\includegraphics[width=3in]{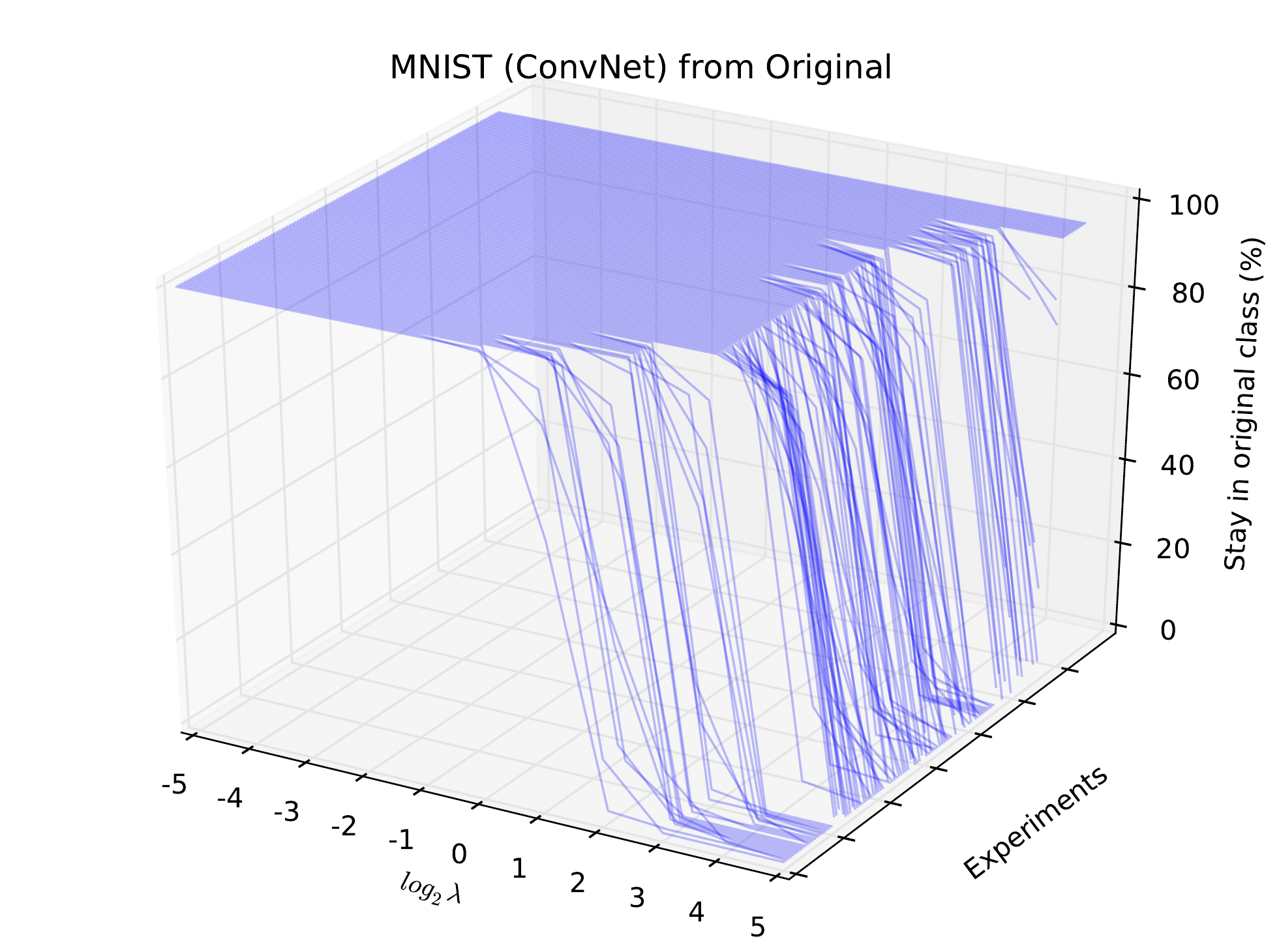}
		\label{fig:mnist_conv_3d_orig_from_orig} }
	\hfil
	\subfloat [] {
		\includegraphics[width=3in]{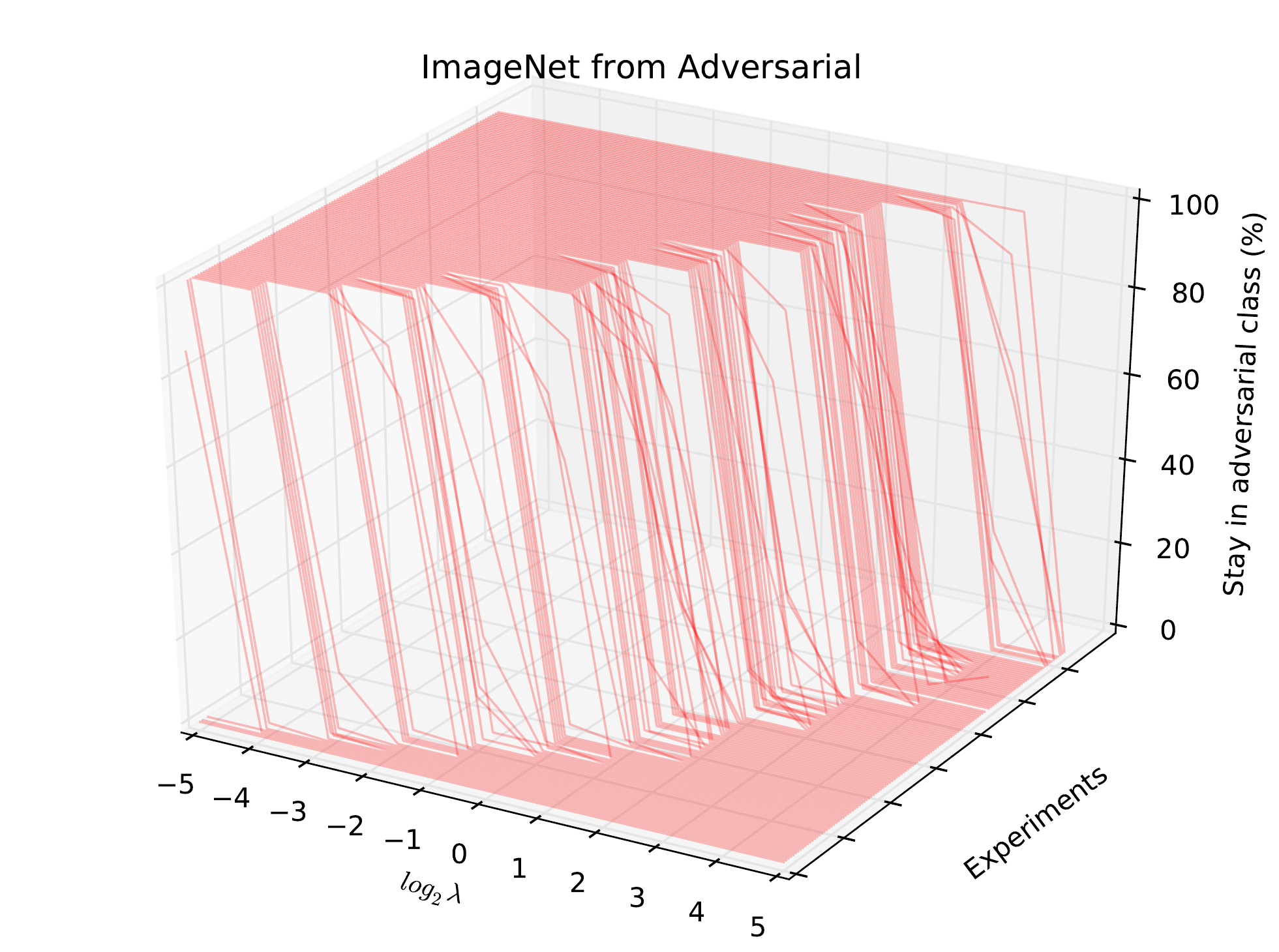}
		\label{fig:3d_adv_from_adv} }
	\subfloat [] {
		\includegraphics[width=3in]{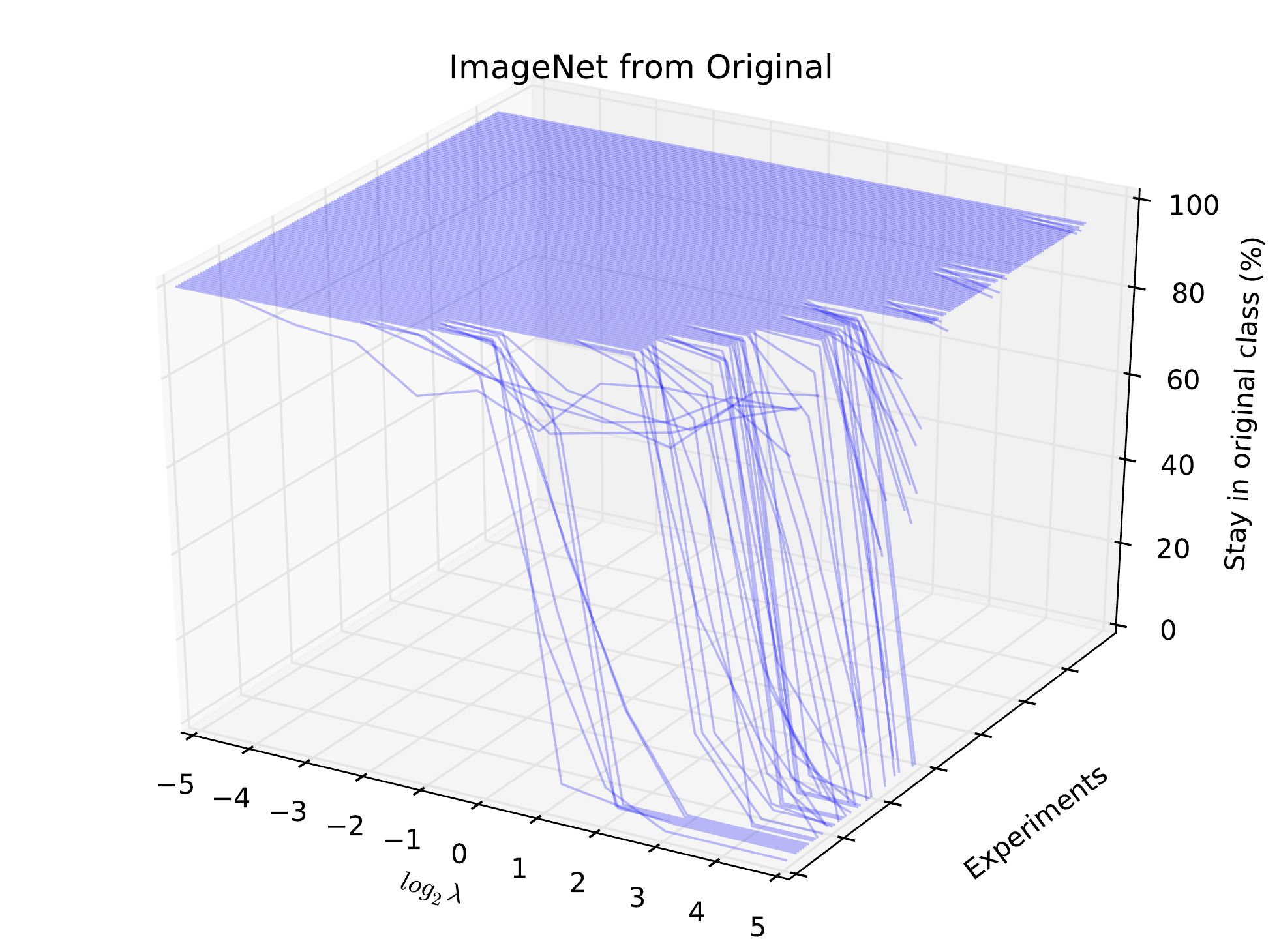}
		\label{fig:3d_orig_from_orig} }
	
	\caption{Adding Gaussian noise to the images. Another view of the probing procedure explained in Section~\ref{sec:methods}. Contrarily to the averaged view of Figure~\ref{fig:all_averages}, here each one of the 125 experiments appears as an independent curve along the \emph{Experiments} axis (their order is arbitrary, chosen to reduce occlusions). Each point of the curve is the fraction of probes (out of a hundred performed) that keeps their class label.}
	\label{fig:all_3d}
\end{figure*}

\begin{figure*}[t!]
	\centering
	\subfloat [MNIST / logistic regression] { 
		\includegraphics[width=3in]{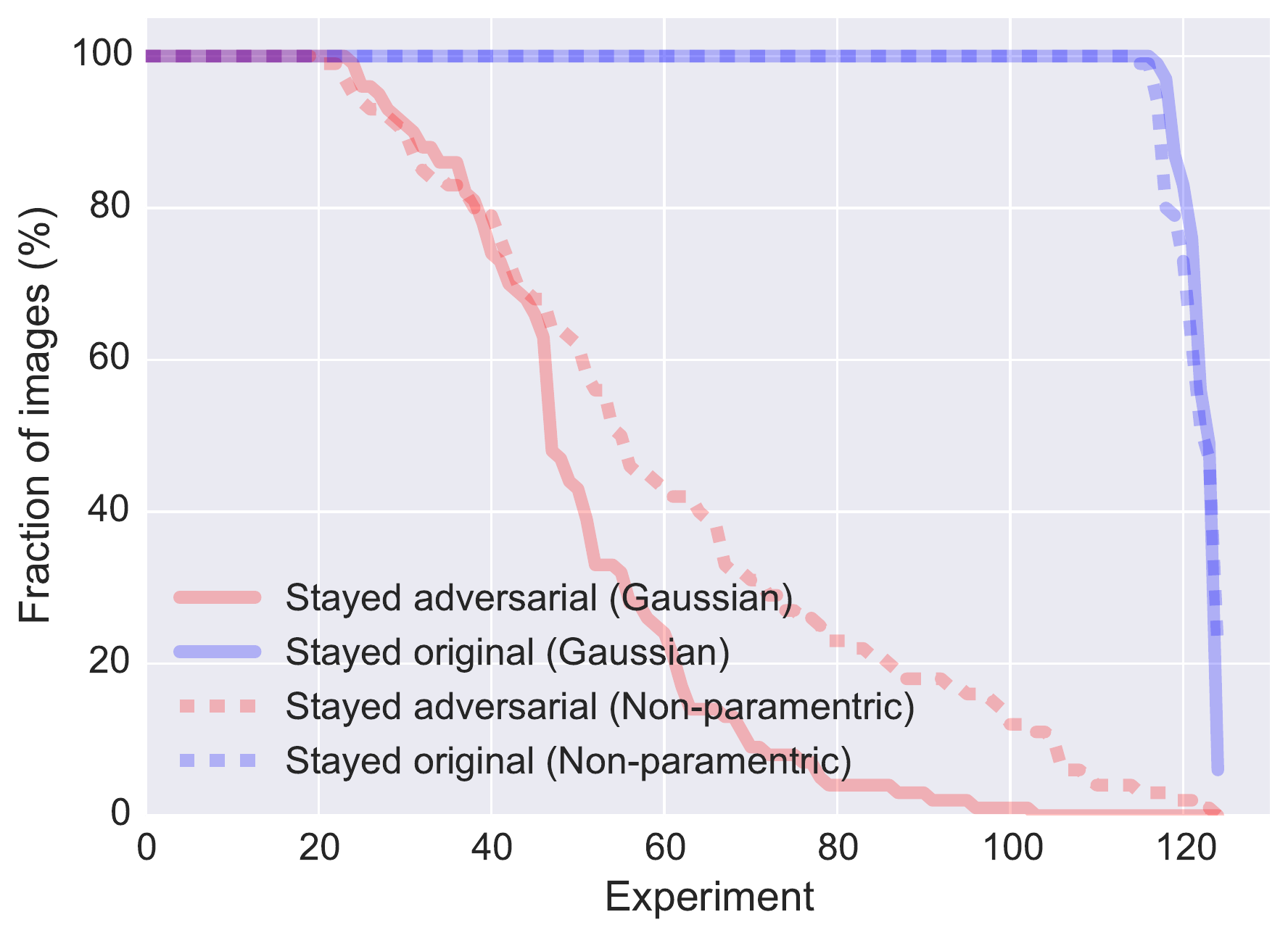}
		\label{fig:mnist_hist} }
	\hfil
	\subfloat [MNIST / convolutional network] {
		\includegraphics[width=3in]{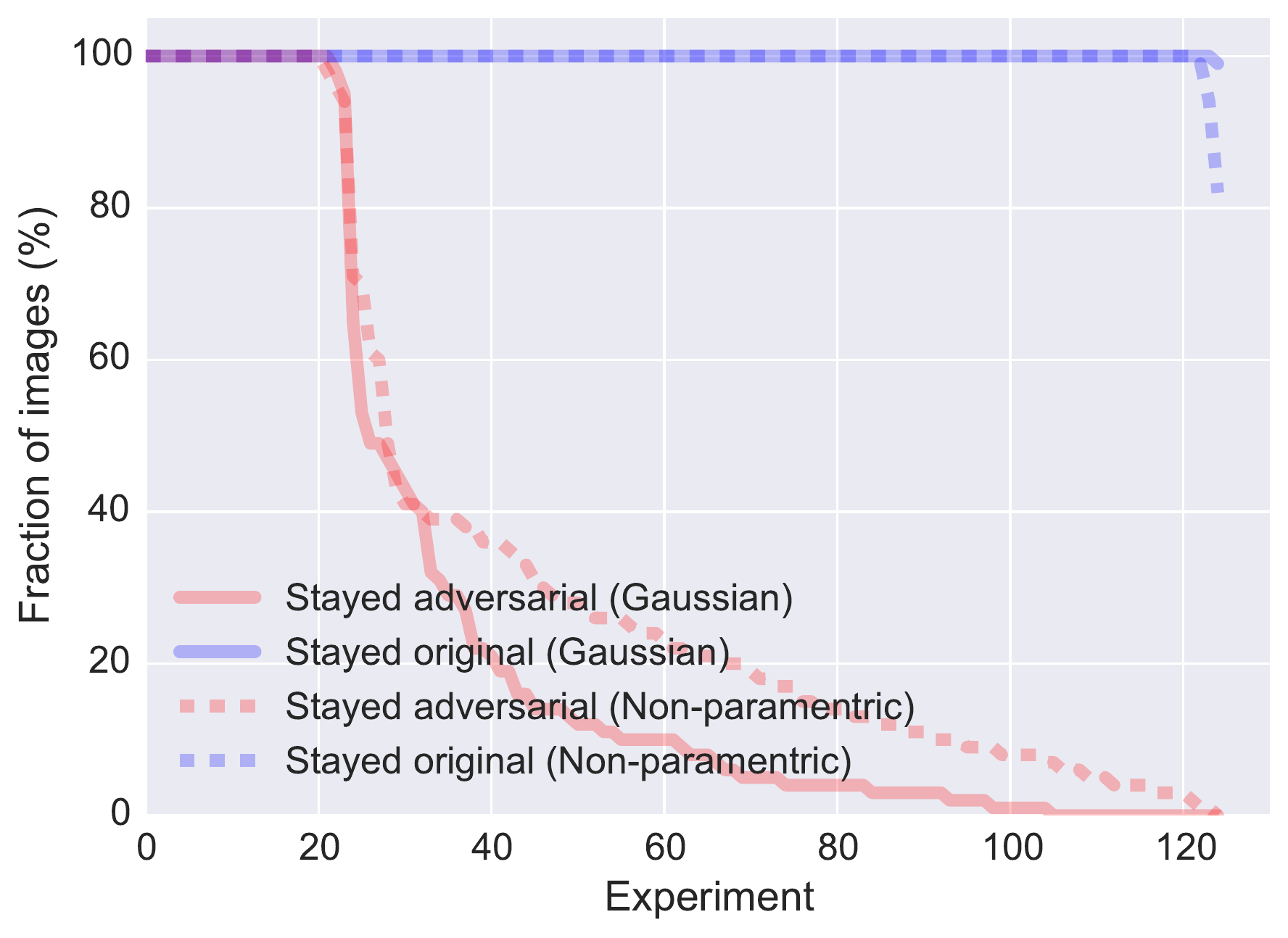}
		\label{fig:mnist_conv_hist} }
	\hfil
	\subfloat [ImageNet / OverFeat] { 
		\includegraphics[width=3in]{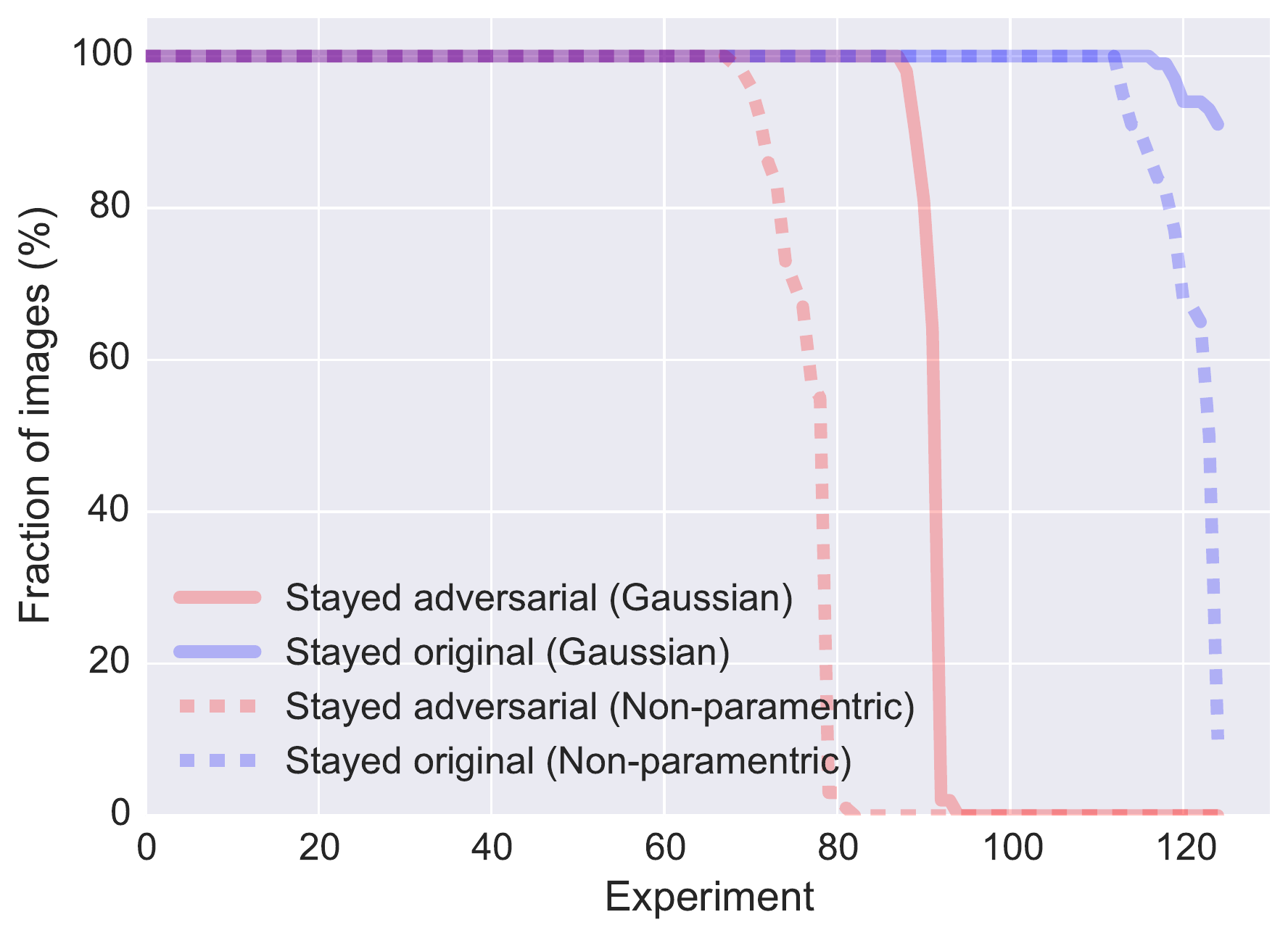}
		\label{fig:hist} }
	
	\caption{For each of the 125 experiments we measure the fraction of the probe images (i.e., departing image + random noise) that stayed in the same class label. Those fractions are then sorted from biggest to lowest along the \emph{Experiments} axis. The area under the curves indicates the entire fraction of probes among all experiments that stayed in the same class.} 
	\label{fig:both_hist}	
\end{figure*}

The results in Figure \ref{fig:all_averages} are strongly averaged, each data point summarizing, for a given level of noise, the result of 125 experiments: the fraction of images that fall in each label for \emph{all} five original class labels, \emph{all} five original samples from each label, and \emph{all} five adversarial class labels. In reality there is a lot of variability that can be better appreciated in Figure~\ref{fig:all_3d}. Here each curve alongside the axis \emph{experiments} represents a \emph{single} choice of original class label, original sample, and adversarial class label, thus there are 125 curves. (The order of the curves along this axis is arbitrary and chosen to minimize occlusions and make the visualization easier). The graphs show that depending on a specific configuration, the label may be very stable and hard to switch (curves that fall later or do not fall at all), or very unstable (curves that fall early). Those 3D graphs also reinforce the point about the stability of the correctly classified original images. 

The results suggest that the classifiers for MNIST are more resilient against adversarial images than ImageNet/OverFeat. Moreover, the shallow MNIST/logistic behaves differently than the deep MNIST/ConvNet, as shown by the the ``falling columns'' in Figure \ref{fig:all_averages}: initially, a small push in MNIST/logistic throws a larger fraction of the adversarial examples back to the correct space. However, at large noise levels, MNIST/logistc saturates with a larger fraction of images still adversarial than MNIST/ConvNet. 


Finally, we wanted to investigate how the nature of the noise added affected the experiments. Recall that our i.i.d. Gaussian noise differs from the original optimized distortion in two important aspects: no spatial correlations, and no important higher-order momenta. To explore the influence of those two aspects, we introduced a noise modeled after the empirical distribution of the distortion pixels. This still ignores spatial correlations, but captures higher-order momenta. The statistics of the distortion pixels are summarized in Table~\ref{table:momenta}, and reveal a distribution that is considerably heavier-tailed than the Gaussians we have employed so far.



Figure~\ref{fig:both_hist} contrasts the effect of this noise modeled non-parametrically after the distortions with the effect of the comparable Gaussian noise ($\lambda=1$). Each point in the curves is one of the 125 experiments, and represents the fraction of the 100 probe images that stays in the same class as the departing --- adversarial or original --- image. The experiments where ordered by this value in each curve (thus the order of the experiments in the curves is not necessarily the same). Here the individual experiments are not important, but the shape of the curves: how early and how quickly they fall. 

For ImageNet, the curves for the non-parametric noise (dotted lines) fall before the curves for the Gaussian noise (continuous line), showing that, indeed, the heavier tailed noise affects the images more, even without the spatial correlation. In addition, all curves fall rather sharply. This shows that in almost all experiments, either all probes stay in the same label as the original, or all of them switch. Few experiments present intermediate results. This rather bimodal behavior was already present in the curves of Figure~\ref{fig:all_3d}.

For MNIST, again, the effect is different: Gaussian and heavy-tail noises behave much more similarly and the curves fall much more smoothly.


\section{Conclusion}

Our in-depth analysis reinforces previous claims found in the literature \cite{goodfellow2014explaining, gu2014towards}: adversarial images are not necessarily isolated, spurious points: many of them inhabit relatively dense regions of the pixel space. This helps to explain why adversarial images tend to stay adversarial across classifiers of different architectures, or trained on different sets\cite{szegedy2013intriguing}: slightly different classification boundaries stay confounded by the dense adversarial regions.

The nature of the noise affects the resilience of both adversarial and original images. The effect is clear in ImageNet/OverFeat, where a Gaussian noise affects the images less than a heavy-tailed noise modeled after the empirical distribution of the distortions used to reach the adversarial images in the first place. An important next step in the exploration, in our view, is to understand the spatial nature of the adversarial distortions, i.e., the role spatial correlations play.

Recent works have attributed susceptibility to adversarial attacks to the linearity in the network \cite{goodfellow2014explaining}, but our experiments suggest the phenomenon may be more complex. A weak, shallow, and relatively more linear classifier (logistic regression), seems no more susceptible to adversarial images than a strong, deep classifier (deep convolutional network), for the same task (MNIST). A strong deep model on a complex task seems to be more susceptible. Are adversarial images an inevitable Achilles' heel of powerful complex classifiers? Speculative analogies with the illusions of the Human Visual System are tempting, but the most honest answer is that we still know too little. Our hope is that this article will keep the conversation about adversarial images ongoing and help further explore those intriguing properties.




\section*{Acknowledgments}
We would like to thank Micael Carvalho for his helpful hints and revision; Jonghoon Jin for open access to his OverFeat wrapper\footnote{\url{https://github.com/jhjin/overfeat-torch}}; and Soumith Chintala, and the rest of the Torch7 team for all of their help on the mailing list. We thank the Brazilian agencies CAPES, CNPq and FAPESP for financial support.

\bibliography{ref}
\bibliographystyle{IEEEtran}
%

\end{document}